\documentclass[runningheads]{llncs}
\usepackage{graphicx}
\usepackage{hyperref}
\usepackage{tikz}
\usepackage{comment}
\usepackage{amsmath,amssymb} % define this before the line numbering.
\usepackage{color}

\usepackage{orcidlink}

\usepackage[accsupp]{axessibility}  % Improves PDF readability for those with disabilities.

\newcommand{\myparagraph}[1]{\smallskip\noindent\textbf{#1}}

\newcommand{\systemname}[1]{\emph{RadioTransformer}}

\begin{document}
% \linenumbers
\pagestyle{headings}
\mainmatter
\def\ECCVSubNumber{7045}  % Insert your submission number here

\title{RadioTransformer: A Cascaded Global-Focal Transformer for Visual Attention--guided Disease Classification} % Replace with your title

\begin{comment}
\titlerunning{ECCV-22 submission ID \ECCVSubNumber} 
\authorrunning{ECCV-22 submission ID \ECCVSubNumber} 
% \author{Anonymous ECCV submission}
% \institute{Paper ID \ECCVSubNumber}
\end{comment}
%******************

% \begin{comment}
\titlerunning{RadioTransformer}
\author{Moinak Bhattacharya
%\inst{1}
\orcidlink{0000-0003-2378-5632} \and
Shubham Jain
%\inst{1}
\orcidlink{0000-0002-4864-6420} \and
Prateek Prasanna
%\inst{1}
\orcidlink{0000-0002-3068-3573}}
\authorrunning{M. Bhattacharya et al.}
\institute{Stony Brook University, Stony Brook, New York, USA
\email{\{moinak.bhattacharya,prateek.prasanna\}@stonybrook.edu}}

%\url{http://www.springer.com/gp/computer-science/lncs} 
% \and
% \email{\{moinak.bhattacharya,prateek.prasanna\}@stonybrook.edu} \and jain@cs.stonybrook.edu}
% \end{comment}

% % CAMERA READY SUBMISSION
% % \begin{comment}
% % If the paper title is too long for the running head, you can set
% % an abbreviated paper title here
% %
% \author{Moinak Bhattacharya\inst{1}\orcidlink{0000-0003-2378-5632} 
% \and
% Shubham Jain\inst{1}%\orcidID{1111-2222-3333-4444} 
% \and
% Prateek Prasanna\inst{1}\orcidlink{0000-0002-3068-3573}
% }
% %
% \authorrunning{M. Bhattacharya et al.}
% \institute{Stony Brook University, Stony Brook NY 11794, USA
% % \and
% % Springer Heidelberg, Tiergartenstr. 17, 69121 Heidelberg, Germany
% % \email{jain@cs.stonybrook.edu}\\
% % \url{http://www.springer.com/gp/computer-science/lncs} \and
% % ABC Institute, Rupert-Karls-University Heidelberg, Heidelberg, Germany\\

% %  \email{\{moinak.bhattacharya,prateek.prasanna\}@stonybrook.edu}

% }
% % \end{comment}
%******************
\maketitle

\begin{abstract}
In this work, we present \systemname~, a novel 
student-teacher transformer framework, that leverages radiologists' gaze patterns and models their visuo-cognitive behavior for disease diagnosis on chest radiographs. 
Domain experts, such as radiologists, rely on visual information for medical image interpretation. On the other hand, deep neural networks have demonstrated significant promise in similar tasks even where visual interpretation is challenging.
Eye-gaze tracking has been used to capture the viewing behavior of domain experts, lending insights into the complexity of visual search. However, deep learning frameworks, even those that rely on attention mechanisms, do not leverage this rich domain information for diagnostic purposes. 
\systemname~ fills this critical gap by learning from radiologists' visual search patterns, encoded as `human visual attention regions' in a cascaded global-focal transformer framework. The overall `global' image characteristics and the more detailed `local' features are captured by the proposed global and focal modules, respectively. We experimentally validate the efficacy of~\systemname~
on 8 datasets involving different disease classification tasks where eye-gaze data is not available during the inference phase. Code: \href{https://github.com/bmi-imaginelab/radiotransformer}{https://github.com/bmi-imaginelab/radiotransformer}
\end{abstract}
\begin{figure}[t]
  \centering
  \includegraphics[width= 0.6\linewidth]{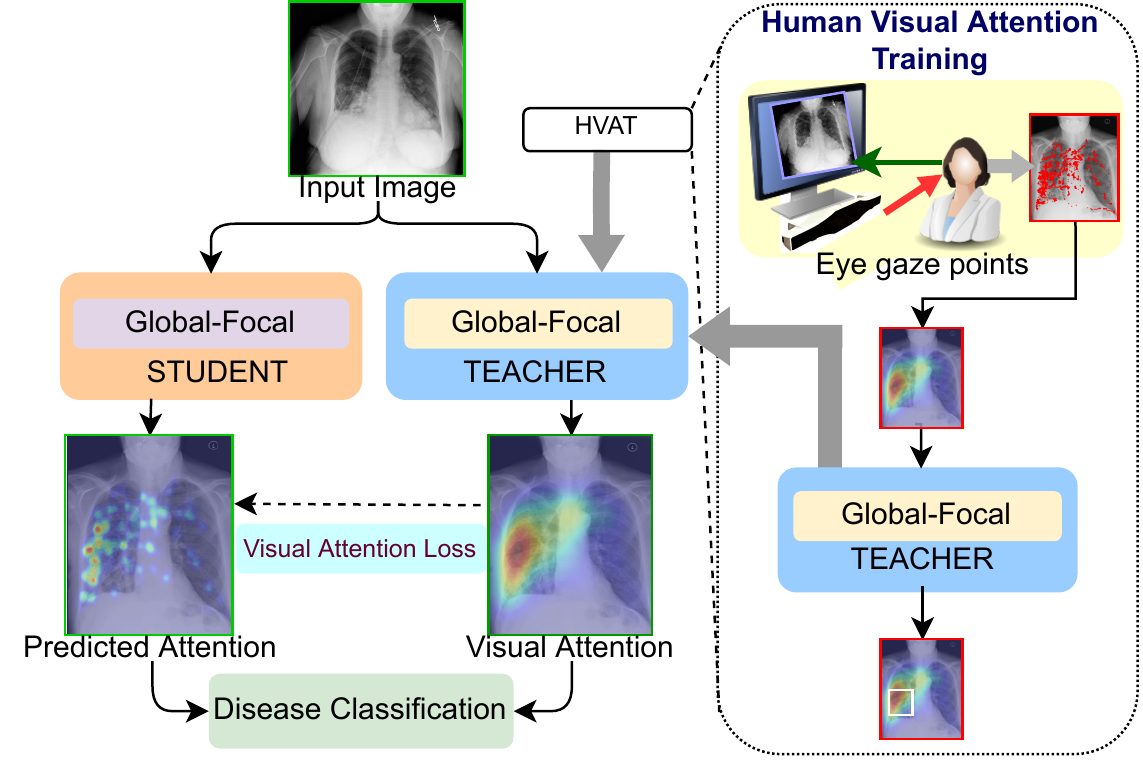}
   \caption{\textbf{Overview of proposed work.} Visual search patterns of radiologists on chest radiographs are used to first train a global-focal teacher network, referred to as \textit{Human Visual Attention Training} (Section~\ref{hva}). This pre-trained teacher network teaches the global-focal student network to learn visual attention 
   using a novel \textit{Visual Attention Loss} (Section~\ref{val}). The student-teacher network is implemented to explicitly integrate radiologist visual attention for improving disease classification on chest radiographs.}
  \label{fig_1}
\end{figure}
\section{Introduction}
\label{sec:intro}

Medical image interpretation and associated diagnosis relies largely on how domain experts study images. 
Radiologists hone their image search skills during years of training on medical images from different domains.
In fact, studies have shown that systematic visual search patterns can lead to improved diagnostic performance~\cite{van2017visual,epp1}.
Current diagnostic and prognostic models, however, are limited to image content semantics such as disease location, annotation, and severity level, and do not take this rich auxiliary domain knowledge into account.
They primarily implement hand-crafted descriptors or deep architectures that learn textural and spatial features of diseases~\cite{cheerla2019deep,prasanna2017radiographic}. 
The spatial dependencies of intra-image disease patterns,
often implicitly interpreted by expert readers, may not be adequately captured via image feature representation learning alone. 

Recent works have utilized transformer-based architectures that leverage attention from radiological scans to provide better diagnosis~\cite{c27,c26}. This is a significant advancement, as the models learn self-attention across image patches to determine diagnostically relevant regions-of-interest. Although these approaches integrate long-range feature dependencies and learn high-level representations, they lack apriori domain knowledge, fundamentally rooted in disease pathophysiology and its manifestation on images. Recently, it has been demonstrated that deep-learning networks can be trained to learn radiologists' attention level and decisions~\cite{mall2019can}. However, it is still unclear how effectively and efficiently such search patterns can be used to improve a model's decision-making ability. 
To address this gap, we propose to leverage domain experts' systematic viewing patterns, as the basis of underlying attention and intention, to guide a deep learning network towards improved disease diagnosis. 

\textbf{Motivation.} The motivation for our approach stems from a) understanding the importance of human visual attention in medical image interpretation, and b) understanding the medical experts' search heuristics in decision-making. Medical image interpretation is a complex process that broadly comprises a global-focal approach involving a) identifying suspicious regions from a global perspective, and b) identifying specific abnormalities with a focal perspective. During the global screening process, a radiologist scans for coarse low-contrast features in which certain textural attributes are analyzed and prospective abnormal regions of interest are identified. In the focal process, the regions of abnormalities are re-examined to determine the severity, type of disease, or reject the assumption of abnormality. For example, while analyzing a chest radiograph for COVID-19, a radiologist skims through the thoracic region at a glance to identify suspicious regions based on intensity variations. This helps in selective identification by eliminating `obviously healthy' regions. The focal feature learning process involves a more critical analysis of the suspicious regions to understand the structural and morphological characteristics of specific regions and their surroundings. This typically involves domain-specific features such as distribution of infiltrates and accumulation of fluid. We use this as a motivation to design \systemname~, a global-focal transformer that integrates a radiologist's visual cognition with the self-attention-based learning of transformers. This improves their class activation regions, leading to a probabilistic score from attention features that correlates highly with human visual attention based diagnosis.

The objective of our work is \textit{to augment the %the fast 
learning 
capabilities of deep networks in a disease diagnosis setting with domain-specific expert viewing patterns in a cognitive-aware manner}.

\textbf{Contributions.} The primary contributions of this work can be summarized as follows:
\begin{enumerate}
    \item A novel \textit{student-teacher based global-focal \textbf{\systemname} architecture}, constituting transformer blocks with shifting windows, is proposed to leverage the radiologists' visual attention in order to improve the diagnostic accuracy. The global module learns high-level coarse representations and the focal module learns low-level granular representations with two-way lateral connections to address the semantic attention gap with smoothed moving average training.
    \item A novel \textit{visual attention loss} (VAL) is proposed to train the student network 
    with the visual attention regions from the teacher network. 
    This loss 
    teaches the student network to focus on regions from teacher-generated visual attention using a weighted combination of attention region overlap and regression of center and boundary points.
\end{enumerate}

Figure~\ref{fig_1} shows an overview of the proposed \systemname~ architecture consisting of the global-focal student-teacher network with a novel Visual Attention Loss. While the underlying concepts of the proposed framework are domain-agnostic, in this work we have validated it on pulmonary and thoracic disease classification on chest radiographs.

\begin{figure}[t]
  \centering
  \includegraphics[width=0.4\linewidth]{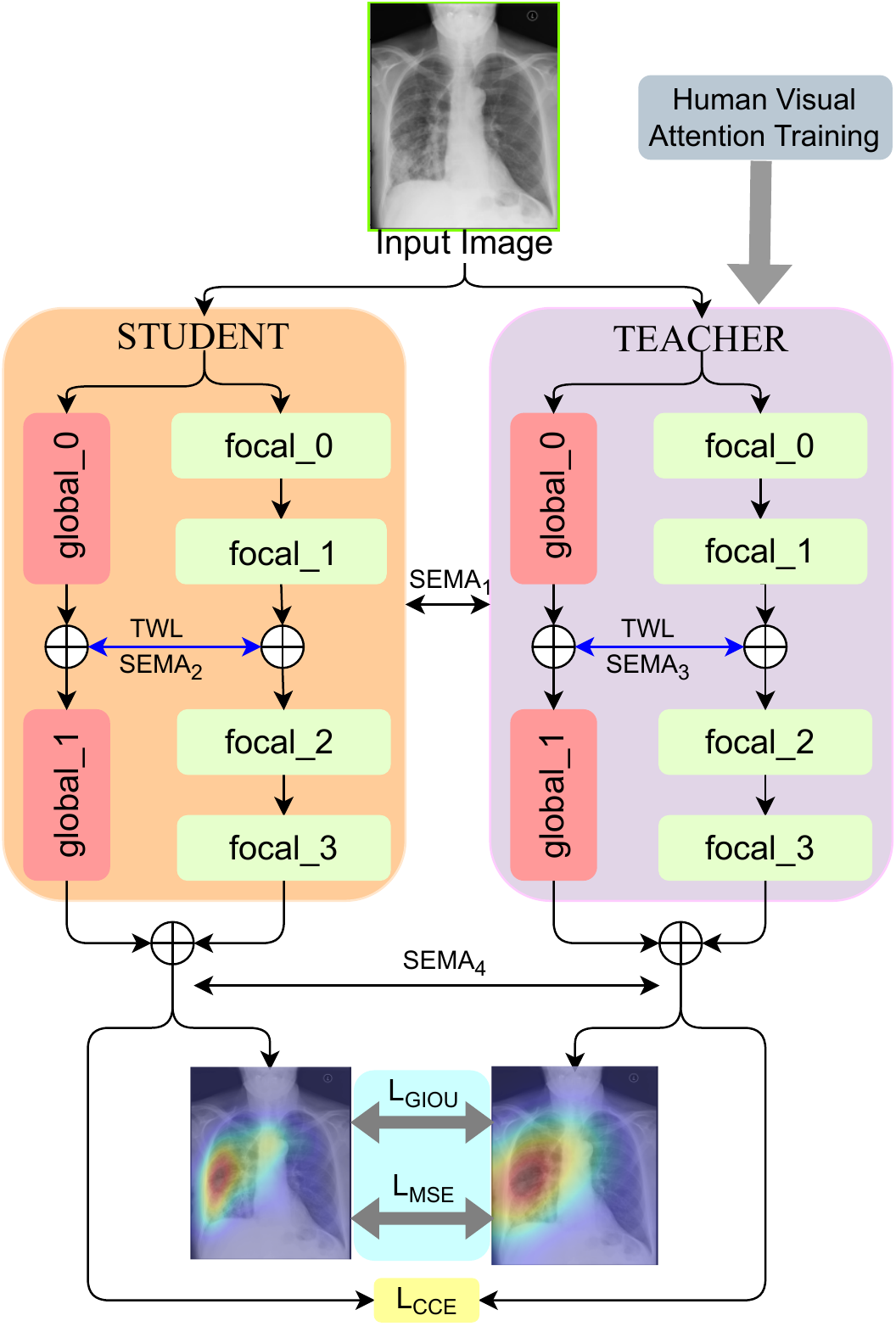}
    \caption{\textbf{Global-Focal Student-Teacher network} implemented using shifting window blocks cascaded in series with TWL connections and layered SEMA.}
    \label{fig_2}
\end{figure} 
\section{Related work}
%\noindent\textbf{Role of Visual Attention in Medical Image Interpretation}
\noindent\textbf{Eye-gaze tracking in Radiology.} Eye-tracking studies have been conducted in radiology to draw insights into the visual diagnosis process~\cite{e5,van2017visual}. Experts' visual search patterns have been studied in various diseases~\cite{e3,e7,e10,e12,e16,e19} to understand their relationship with the diagnostic performance of radiologists~\cite{e25,e26,e27}. Clinical error in diagnostic interpretation has often been attributed to reader fatigue and strain, which has been extensively validated via eye-tracking studies~\cite{e21,e22,e23,e24}. Variations in cognition and perceptual patterns while viewing images can cause the same image being interpreted differently by different experts. This has led to a few studies displaying eye-positions from experts as a visual aid to improve diagnostic performance of novice readers~\cite{epp1,epp2}. The dependence of diagnostic decisions on visual search patterns presents a unique opportunity to integrate this rich auxiliary domain information in computer-aided diagnosis systems.

\textbf{Visual attention--driven learning.}
In the context of image interpretation, visual attention refers to the cognitive operations that direct an observer's attention to specific regions in an image. We represent visual attention as saliency maps constructed by tracking users' eye movements. Eye-gaze~\cite{gaze_review} has been used in several computer vision\cite{cv_gaze_review_1,cv_gaze_review_2} studies for head-pose estimation, human-computer interaction, driver vigilance monitoring, etc. Human eyes tend to focus on visual features, such as corners~\cite{va11}, luminance~\cite{va12}, visual onsets\cite{va_1,va_2}, dynamic events~\cite{va13,va14}, color, intensity, and orientation\cite{va15,va16,va17}. Image perception, in general, is hence tightly coupled with visual attention of the observer. Several methods, involving gaze analysis, have been proposed for tasks such as object detection~\cite{va20,va21,va22}, image segmentation~\cite{va18,va19}, object referring~\cite{va2}, action recognition~\cite{va23,va24,va4,va6}, and action localization~\cite{va25}. Other specialized methods use visual attention for goal-oriented localization~\cite{va1} and egocentric activity recognition~\cite{va5}. A recent work incorporated sonographer knowledge in the form of gaze tracking data on ultrasounds to enhance anatomy classification tasks~\cite{va3}. In another study~\cite{interesting_study}, Convolutional Neural Networks (CNN) trained on eye tracking data were shown to be equivalent to the ones trained on manually annotated masks for the task of tumor segmentation. 

Despite evidence of the importance of expert gaze patterns in improving image interpretation, their role in machine-learning driven disease classification in radiology, is still under explored. The interpretation of radiology images is a complex task, requiring specialized viewing patterns unlike the more general visual attention in other tasks. For example, determining whether a lesion is cancerous or not involves the following hierarchical steps: a) detecting the presence of a lesion, b) recognizing whether it is pathologic, c) determining the type, and finally, d) providing a diagnosis. These sequential analysis patterns, to some extent, are captured by the visual search patterns which are not leveraged by machine learning models. To bridge this gap, our proposed work uses the visual attention knowledge from radiologists to train a transformer-based model for improving disease classification on chest radiographs.

\textbf{Disease classification on chest radiographs.}
Reliable classification of cardiothoracic and pulmonary diseases on chest radiographs is a crucial task in Radiology, owing to the high morbidity and mortality resulting from such abnormalities.  
Several methods have been proposed to address this, of which the most prominent baselines, ChexNet~\cite{c11}, and CheXNext~\cite{c12}, use a Densenet-121~\cite{d} backbone. Attention-based models such as $A^3Net$~\cite{c21}, and DuaLAnet\cite{c22}, have also been proposed for this diagnostic task. CheXGCN~\cite{c18} and SSGE\cite{c19} are Graph Convolutional Network (GCN)--based methods; the latter proposes a student-teacher based SSL method. 
More recently, attempts have been made to develop methods for diagnosis and prognosis of COVID-19 from chest radiographs. Most of these methods~\cite{c1,c2,c16,c14,c4} use backbones of deep convolution neural network for COVID-19 prediction. Although, CNN-based methods have achieved tremendous success through generic feature extraction strategies, these architectures often fail to comprehensively encode spatial features from a biological viewpoint\cite{c30_x}.

To address this limitation, transformer-based approaches, such as vision transformers~\cite{vit}, have been proposed. The self-attention mechanism in transformers integrates global information by encoding the relative locations of the patches. Few recent works have proposed vision transformers for COVID-19 prediction task~\cite{c27,c26}. However, the efficacy of shifting window based~\cite{swin} transformer architectures has not been evaluated in this domain. These recent methods compute self-attention among patches within local windows. As an example, Swin-UNet~\cite{swin_unet} implements swin transformer blocks for medical image segmentation. These blocks are well suited to characterize intra-image disease heterogeneity, a very crucial factor affecting diagnosis and patient prognosis. This motivates our choice of using shifting window blocks in the proposed global-focal network. 

\begin{figure*}[t]
  \centering
  \includegraphics[width=1.0\linewidth]{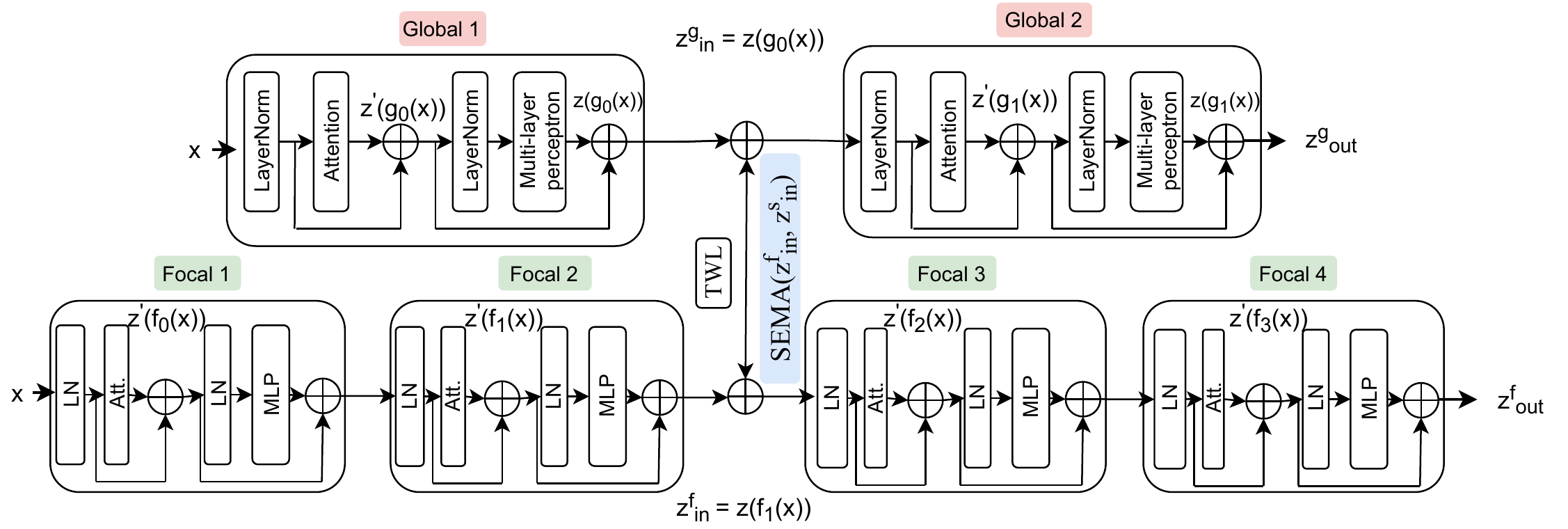}
   \caption{\textbf{Overall global-focal network.} There are two global networks, and four focal networks connected in parallel inside a Student or Teacher network. The components of the global and focal network are similar, where LN: Layer Normalization, Att.: Attention, MLP: Multi-layer Perceptron. The output of Focal 2 is connected with Global 1 with a TWL connection with SEMA applied to it.}
   \label{fig_4}
\end{figure*}

\section{Proposed methodology}
\label{propmethod}
\label{proposed_methodology}
Figure~\ref{fig_2} presents an overview of the end-to-end framework of the proposed \systemname~ global-focal student-teacher network. This comprises two parallel architectures, a student and a teacher model.  Both student and teacher networks have global and focal network components. Four focal blocks  in each model are cascaded with two global blocks in parallel. 
The global and focal blocks are connected via a two-way lateral (TWL) connection~\cite{christoph2016spatiotemporal,feichtenhofer2019slowfast,lin2017feature} with smoothed exponential moving average (SEMA). 
SEMA regulates the attention features shared between the global and focal blocks to bridge the attention gap caused by different learning scales across these networks.

The teacher model is trained with human visual attention obtained from visual search patterns of radiologists. The student model learns from the teacher network using VAL and a classification loss.
There are two TWL connections between the teacher and student models coupled with layered SEMA.
The proposed architecture is explained in the following subsections.

\subsection{Global-focal architecture}
\label{gfst}
Global-focal networks can be described as a single-stream architecture where the two components operate in parallel. The global network consists of two and the focal network consists of four shifting-window transformer blocks (Figure~\ref{fig_4}). This draws its analogy from the pathways that involve the Parvo, Magno, and Konio ganglion cells~\cite{pm_1,pm_2}. 
The focal network is inspired by the functioning of slow responding Parvo cells (in the `what' pathway), and the global network is inspired by the fast Magno cells (in the `where' pathway).

\myparagraph{Global-focal network.}
\label{gf}
The teacher and student networks are variants of global-focal architecture. The primary idea of the global-focal architecture is to pseudo-replicate learning of attention in a detailed shifting window fashion as shown in Supplementary Figure 1. The focal and global layers are represented as $f_i$ and $g_j$, respectively, where $i \in \{0, 1, 2, 3\}$ and $j \in \{0, 1\}$. 

\textbf{Focal network.}\label{f} The focal network is implemented to learn high contrast and focal information from shifting the windows incrementally on four blocks that are cascaded in a series. The first block of the focal network has multi-layer perceptron head, $h^{mlp}_{f_0}=64$, attention head, $h^{att}_{f_0}=2$, and shift size, $s_{f_0}=0$. The second, third and fourth blocks operate with incremental shifting window size $s_{f_i} = \{1, 2, 3\}$, $h^{att}_{f_i} = \{4, 4, 8\}$ and $h^{mlp}_{f_i} = \{128, 128, 256\}$, where $i \in \{1, 2, 3\}$. 

\textbf{Global network.}\label{g} The global network consists of two shifting-window blocks cascaded in series. The motivation for implementing global network is to learn low contrast global information from two incremental shift sizes. The first block in the global network has a shift size $s_{g_0}=0$ and the second block has a shift size $s_{g_1}=1$. The multi-layer perceptron head of the global network is incremental and can be represented as $h^{mlp}_{g_j}=\{128, 256\}$. The attention head of the global network is incremental and can be represented as $h^{att}_{g_j}=\{4, 8\}$, where $j \in \{0, 1\}$.

\textbf{TWL connections.}
\label{gf_skip}
TWL connections between global and focal architectures are introduced to address the inherent semantic attention averaging between the two. The TWL connections are established between layers \{$f_1$, $g_0$\} and \{$f_3$, $g_1$\}. 
These constitute weighted addition of the outputs from the aforementioned layers coupled with SEMA on the weighted addition outputs.
This can be represented as,\\
\begin{equation}
    z^{gf}_p = \lambda_{p_1}^{gf} . g_p(x) + \lambda_{p_2}^{gf} . f_p(x)
\end{equation}
where, $\lambda^{gf}_{p_1}$ and $\lambda^{gf}_{p_2}$ are the hyper-parameters for weighted addition of the outputs from the global-focal networks represented as $gf$. $z(g_p(.))$ is the output from the global network and $z(f_p(.))$ is the output from the focal network,  $p \in \{in, out\}$ where \textit{in} is the intermediate, and \textit{out} is the final output. $\{z^f_{in}, z^g_{in}\}:\{z(f_{in}(.)), z(g_{in}(.))\}$ are the outputs from the intermediate layers of the focal and global networks, respectively. $\{z^f_{out}, z^g_{out}\}:\{z(f_{out}(.)), z(g_{out}(.))\}$ are the final outputs from the focal and global networks, respectively. This is shown in Figure~\ref{fig_4}. The smoothed moving average $s_v$ is given by,
\begin{equation}
     s_{v_p}(z_p^{gf}) = \hat{\delta}^{gf}_p . s_{v'_p}(z^{gf}_p) + (1 - \hat{\delta}^{gf}_p) . v_p(z^{gf}_p)
\end{equation}
where $s_{v_p}$ is the smoothed-value of the current variable $v$ in the current iteration for different $p$, and $s_{v'}$ is the smoothed-value of the variable from the previous iteration for a different $p$. $\hat{\delta}^{gf}_p$ is the smoothing decay hyperparameter of the global-focal TWL connection. This is represented as $\hat{\delta}^{gf}_p = 1 - \frac{1}{N}$, where $N$ is the number of samples in the current iteration.

\myparagraph{Student-teacher network.}
\label{st}
A student-teacher network is proposed in this work. 
The teacher network learns visual attention patterns only from radiologist's eye gaze maps, while the student learns more specific disease attributes directly from the medical images by leveraging attention information provided by the teacher. Generally, the visual attention maps from radiologists can be noisy and may exhibit 
variability. Incorporating this variability in addition to distinct disease patterns is not feasible in single-stream architectures.
Hence, we need a student-teacher learning framework so that the student can learn this soft information from the teacher. 
Also, the student-teacher network reduces the complexity of training a single network with the visual attention maps and further fine-tuning for downstream tasks.
Here, the model is compressed with just the teacher trained with the visual attention maps. 

\textbf{Teacher network.}
\label{t} 
The teacher network is a cascaded global-focal learning network with two global and four local blocks connected in parallel, represented as:
\begin{equation}
    z^t_{in} = \lambda_{t_1}^{l_0} . g^t_0(x^t) + \lambda_{t_2}^{l_0} . f^t_1(f^t_0(x^t))
\end{equation}
\begin{equation}
    z^t_{out} = \lambda_{t_1}^{l_1} . g^t_1(z^t_{in}) + \lambda_{t_2}^{l_1} . f^t_3(f^t_2(z^t_{in}))
\end{equation}
where $x^t$ is the input to the teacher network, which is subject to hard augmentation techniques with stateless high-value intervals of brightness, contrast, hue, and saturation. 
$z^t_{in}$ is the intermediate output of the teacher network with $\{\lambda^{l_0}_{t_1}$,  $\lambda^{l_0}_{t_2}\}$, and $\{\lambda^{l_1}_{t_1}$,  $\lambda^{l_1}_{t_2}\}$ as the hyperparameters for weighted addition of the intermediate and final outputs from global and focal blocks, respectively.

\textbf{Student network.}
\label{s}
The input to the student network is softly augmented with stateless relatively low-value intervals of brightness, contrast, hue, and saturation as compared to the teacher network. The student predicts probability values of the disease classes along with an attention region. This attention region is subjected to 
% a self-supervised loss 
VAL, described in Section~\ref{val},
with the output of the attention region from the teacher network. The student network can be represented as
\begin{equation}
    z^s_{in} = \lambda_{s_1}^{l_0} . g^s_0(x^s) + \lambda_{s_2}^{l_0} . f^s_1(f^s_0(x^s))
\end{equation}
\begin{equation}
    z^s_{out} = \lambda_{s_1}^{l_1} . g^s_1(z^s_{in}) + \lambda_{s_2}^{l_1} . f^s_3(f^s_2(z^s_{in}))
\end{equation}
where $x^s$ is the input to the student network.
$z^s_{in}$ is the intermediate output of the student network with $\{\lambda^{l_0}_{s_1}$,  $\lambda^{l_0}_{s_2}\}$, and \{$\lambda^{l_1}_{s_1}$,  $\lambda^{l_1}_{s_2}$\} as the hyperparameters for weighted addition of the intermediate and final outputs from the global and focal blocks of the student network, respectively.

\textbf{TWL connections.}
\label{st_skip}
TWL connections between student and teacher architectures are introduced between layers \{$f_{in}$, $g_{in}$\} and \{$f_{out}$, $g_{out}$\}. The weighted addition of the outputs from the aforementioned layers are coupled with SEMA. This is represented as:

\begin{equation}
    z^{st}_{in} = \lambda_{in_1}^s . z^s_{in} + \lambda_{in_2}^t . z^t_{in}
\end{equation}
\begin{equation}
     s_v(z^{st}_{in}) = \hat{\delta}^{st}_{in} . s_{v'}(z^{st}_{in}) + (1 - \hat{\delta}^{st}_{in}) . v(z^{st}_{in})
\end{equation}
where $z^{st}_{in}$ is the output from the intermediate TWL connection of student-teacher network and $s_v$ is the SEMA from this layer.
\begin{equation}
    z^{st}_{out} = \lambda_{out_1}^s . z^s_{out} + \lambda_{out_2}^t . z^t_{out}
\end{equation}
\begin{equation}
     s_v(z^{st}_{out}) = \hat{\delta}^{st}_{out} . s_{v'}(z^{st}_{out}) + (1 - \hat{\delta}^{st}_{out}) . v(z^{st}_{out})
\end{equation}
where $z^{st}_{out}$ is the output from the final layer of student-teacher network and $\{s_v(z^{st}_{in}), s_v(z^{st}_{out})\}$ are the $\{SEMA_1, SEMA_4\}$, as shown in Figure~\ref{fig_2}. Also, $\{SEMA_2, SEMA_3\}$ are the SEMAs for the intermediate layers of the student global-focal, and teacher global-focal network. The augmentation strategies are explained in the Supplementary section.

\begin{figure}[t]
  \centering
  \includegraphics[width=0.7\linewidth]{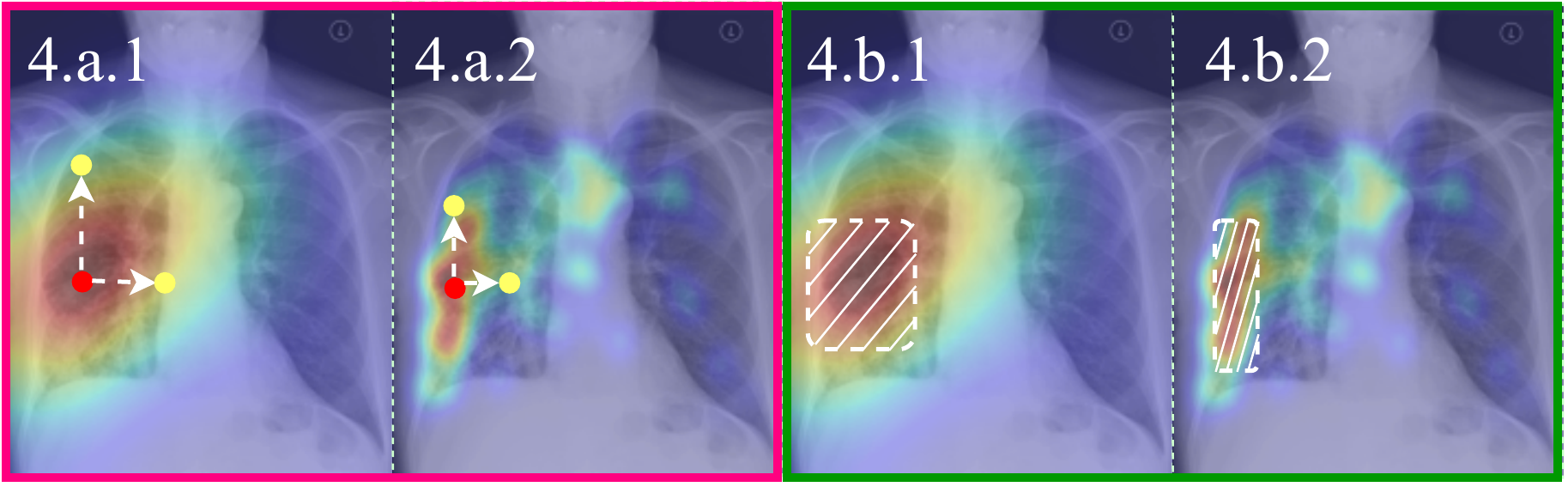}
   \caption{\textbf{Visual Attention Loss, $\mathcal{L}_{VAL}$.} 4.a.* illustrates the computation of $\mathcal{L}_{MSE}$, where the red dot is the center point of the attention region, and the yellow dots are the height, and width. 4.b.* shows $\mathcal{L}_{GIoU}$, where the attention region overlap is shown with dashed boxes. \{4.a.1, 4.b.1\} are the predicted attention regions, and \{4.a.2, 4.b.2\} are the human visual attention regions.
}
   \label{fig_5}
\end{figure}

\subsection{Visual attention loss}
\label{val}
The visual attention regions are obtained from the teacher network and the predicted attention regions are obtained from the student network. We propose a novel % self-supervised 
visual attention loss (VAL) function to train the student network. VAL includes a GIoU and a MSE loss, as shown in Figure~\ref{fig_5}. We use a hyperparameter $\lambda_{l_i} \in \mathbb{R}^+$ to induce weights in the losses with $i \in  \{1, 2\}$.

\begin{equation}
    \mathcal{L}_{GIoU} = 1 - \bigg\{ \dfrac{|(\mathcal{A}_{pred} \cap {\mathcal{A}}_{hva})|}{|(\mathcal{A}_{pred} \cup {\mathcal{A}}_{hva})|} - \dfrac{|C \setminus (\mathcal{A}_{pred} \cup \mathcal{A}_{hva})|}{|C|} \bigg\}
\end{equation}
where ${\mathcal{A}}_{hva}$ is the visual attention region predicted from the teacher network and ${\mathcal{A}}_{pred}$ is the attention region predicted from the student network. $C$ is the smallest convex hull of ${\mathcal{A}}_{hva}$ and ${\mathcal{A}}_{pred}$. The regression loss between  the predicted keypoints and keypoints from visual attention is represented as 
\begin{equation}
   \mathcal{L}_{MSE} = \dfrac{1}{n} \sum_{k=1}^{n} \| (\mathcal{K}_{c_x, c_y, h, w})_k - (\hat{\mathcal{K}}_{c_x, c_y, h, w})_k \|^2_2 
\end{equation}
where $\{c_x, c_y\}$ are the center points and $\{h, w\}$ are height, and width of the attention region. $\mathcal{K}_{(.)}$ is the keypoint of ${\mathcal{A}}_{pred}$. $\hat{\mathcal{K}}_{(.)}$ is the keypoint of ${\mathcal{A}}_{hva}$. $n$ is the number of samples in a particular batch. The final loss is calculated as:
\begin{equation}
   \mathcal{L}_{VAL} = \lambda_{l_1}.\mathcal{L}_{GIoU} + \lambda_{l_2}.\mathcal{L}_{MSE}
\end{equation}
where $\mathcal{L}_{VAL}$ is the proposed VAL and $\{\lambda_{l_1}, \lambda_{l_2}\}$ are the hyperparameters used for weighted addition of the two losses.

\begin{figure}[t]
  \centering
  \includegraphics[width=0.7\linewidth]{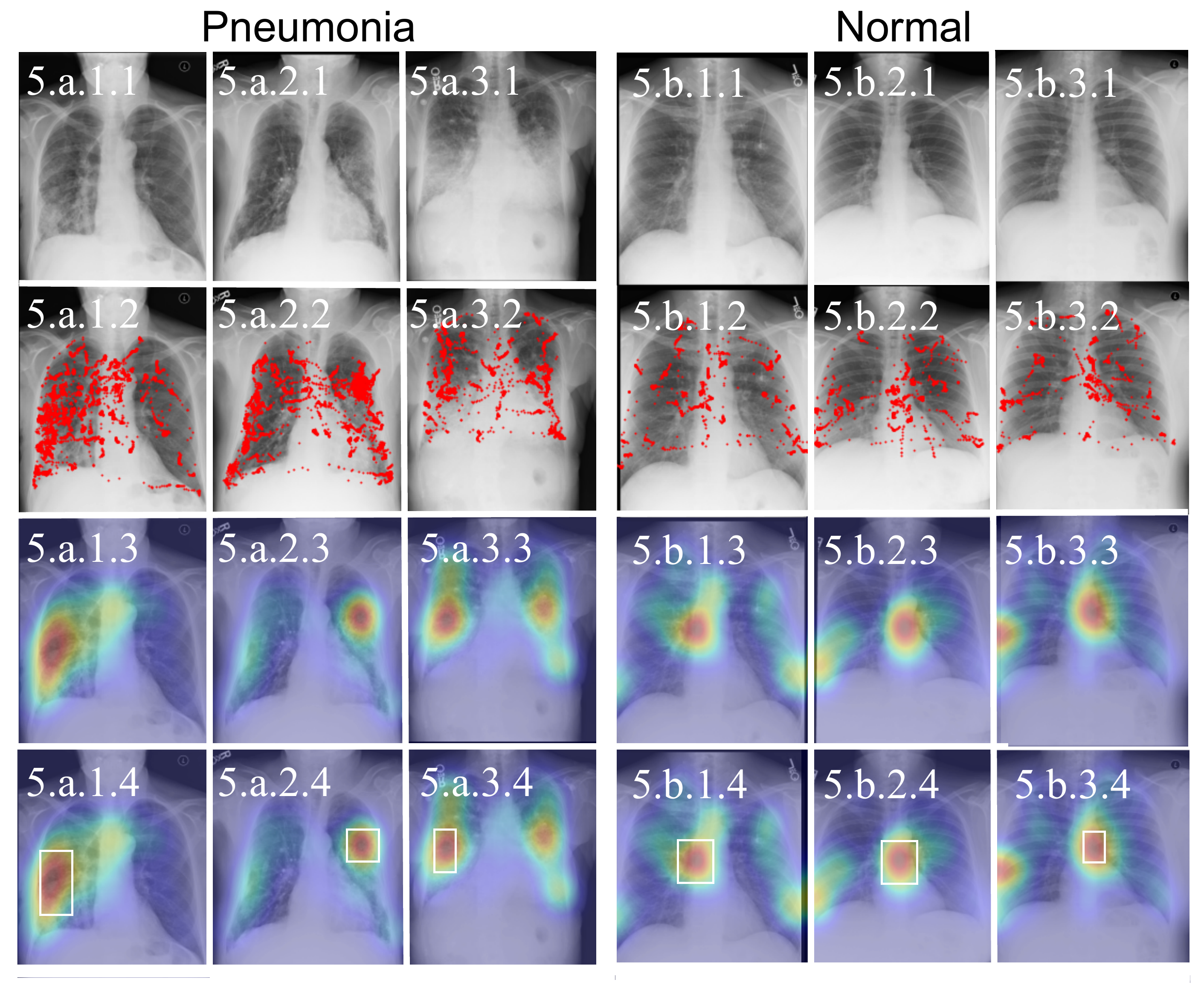}
    \caption{\textbf{Human Visual Attention}. 5.a.*.* series represents Pneumonia and 5.b.*.* series represents normal examples from \cite{e4}. Chest X-Rays from patient are shown in the first row as 5.*.*.1, the raw eye-gaze points from radiologists is shown in the second row as 5.*.*.2, human visual attention maps are shown in the third row as 5.*.*.3, and the corresponding bounding boxes are shown in the fourth row as 5.*.*.4}.
    \label{fig_3}
    \vspace{-25pt}
\end{figure}

\subsection{Human visual attention}
\label{hva}
\textbf{Pre-processing.}
In this subsection, we discuss the methodology for extracting visual search patterns from eye-tracking data and generating visual attention maps of radiologists. The eye-tracking data~\cite{e4} consists primarily of  a) raw eye-gaze information (as shown in Figure~\ref{fig_3}.*.*.2), and b) fixations information, captured from radiologists while they are analyzing chest radiographs in a single-screen setting.
The eye-gaze points are reflective of the diagnostic search patterns. The cumulative attention regions, represented as heatmaps (Figure~\ref{fig_3}.*.*.3), are human attention regions reflective of diagnostically important areas. A multi-dimensional Gaussian filter with standard deviation, $\sigma = 64$, is used to generate these attention heatmaps. Contours from these attention heatmaps are selected with a thresholding value of $\lambda = 140$ and, subsequently, bounding boxes are generated from the contour with the largest area, as shown in Figure ~\ref{fig_3}.*.*.4.

\textbf{Human visual attention training (HVAT).}  
Next, the teacher network is trained with the eye-tracking data from~\cite{e4}.
The teacher network has a classification head to provide an output probability value and a detection head to output key points. The probability value is a $1 \times n$ vector, where $n$ represents the number of different types of disease labels. The key-points output is $\{x_c,y_c, h, w\}$, where $(x_c, y_c)$ are the x and y coordinates of the center, and $(h, w)$ are the height and width respectively. Also, Categorical Crossentropy loss is used for classification, and weighted addition of Generalized Intersection-with-Union (GIoU) loss\cite{giou} and Mean Squared Error (MSE) loss for detection.
\section{Datasets and environment}
\noindent\textbf{Datasets.} 
The proposed architecture is evaluated on eight different datasets consisting of two pneumonia classification, four COVID-19 classification (TCIA-SBU~\cite{sbu,clark2013cancer}, and MIDRC \cite{midrc,midrc_1,clark2013cancer} only for testing)
, and two thoracic disease classification cohorts. Further dataset details are provided in the Supplementary section. 
The datasets along with the train-validation-test splits are shown in Table~\ref{tab_1}.

\begin{table}[t]
\begin{center}
\caption{Train-Validation-Test splits used across all experiments}
\scalebox{0.7}{
\begin{tabular}{*{9}{c}}
\hline
Name & Cell & RSNA & SIIM & Rad & NIH & VBD & MIDRC & SBU \\
\hline
Train & 4200 & 21158 & 4433 & 14815 & 77871 & 47539 & - & - \\
Valid & 1032 & 3022 & 633 & 2116 & 17304 & 6791 & - & -  \\
Test & 624 & 6045 & 1266 & 4233 & 25596 & 13582 & 1241 & 14220 \\
\hline
\end{tabular}
}

\label{tab_1}
\end{center}
\end{table}

% \begin{table}
% \begin{center}
% \caption{Font sizes of headings. Table captions should always be
% positioned {\it above} the tables. The final sentence of a table
% caption should end without a full stop}
% \label{table:headings}
% \begin{tabular}{lll}
% \hline\noalign{\smallskip}
% Heading level & Example & Font size and style\\
% \noalign{\smallskip}
% \hline
% \noalign{\smallskip}
% Title (centered)  & {\Large \bf Lecture Notes \dots} & 14 point, bold\\
% 1st-level heading & {\large \bf 1 Introduction} & 12 point, bold\\
% 2nd-level heading & {\bf 2.1 Printing Area} & 10 point, bold\\
% 3rd-level heading & {\bf Headings.} Text follows \dots & 10 point, bold
% \\
% 4th-level heading & {\it Remark.} Text follows \dots & 10 point,
% italic\\
% \hline
% \end{tabular}
% \end{center}
% \end{table}

\textbf{Environment.} All experiments were performed on the Google Cloud Platform in a compute node with 2 vCPUs, 16 GB RAM, and 20 GB disk memory. The baselines and proposed architectures were trained on a cloud TPU of either type v2-8 or v3-8 with version 2.8.0. All implementations are in TensorFlow~\cite{abadi2016tensorflow} and Keras~
\cite{chollet2015keras} v2.8.0.
\section{Experiments and results}
\label{exp_res}
\noindent\textbf{Implementation.} During HVAT, the teacher network is trained on eye-gaze data from~\cite{karargyriseye,goldberger2000physiobank} which contains radiologist eye-gaze points on 1083 chest x-rays from the MIMIC-CXR dataset~\cite{johnson2019mimic,johnson2020mimic,goldberger2000physiobank} (details in subsection~\ref{hva}). All the images are resized to $256\times256$ pixels. The output of the teacher network is a $1 \times 3$ vector of probability values 
and a $1 \times 4$ vector of keypoints. All the baseline models are trained with images uniformly resized to $256\times256$ pixels. They are trained with Adam optimizer with a batch size of 64 for 50 epochs. The initial learning rate (LR) is set to $1 \times 10^{\text{-}2}$. The LR is scheduled with an exponential LR scheduler with decay steps~=~$10^5$ and decay rate~=~0.2. There is an early stopping criteria with patience~=~20 with the task to minimize the validation loss. The proposed \systemname~ architecture follows the same training standards. 

\begin{table*}
%\tiny
\begin{center}
\caption{\textbf{Quantitative Comparison.} F1($\uparrow$) and AUC($\uparrow$) are reported for the baselines and \systemname~(RadT)}
\scalebox{0.61}{
\begin{tabular}{*{13}{c}|cccc}
\hline
Classification$\rightarrow$ & \multicolumn{4}{c}{Pneumonia} & \multicolumn{4}{c}{COVID-19} & \multicolumn{4}{c}{14-Thoracic} & \multicolumn{4}{c}{COVID-19 (Test)}\\
\hline
Dataset$\rightarrow$ & \multicolumn{2}{c}{Cell\cite{cell}} & \multicolumn{2}{c}{RSNA\cite{rsna}} & \multicolumn{2}{c}{SIIM\cite{siim}} & \multicolumn{2}{c}{Rad\cite{rad,rad_1}} & \multicolumn{2}{c}{NIH\cite{nih}} & \multicolumn{2}{c|}{VBD\cite{vbd}} & \multicolumn{2}{c}{MIDRC\cite{midrc_1,midrc}} & \multicolumn{2}{c}{SBU\cite{sbu,clark2013cancer}}\\
\hline
Architectures$\downarrow$ & F1 & AUC & F1 & AUC & F1 & AUC & F1 & AUC & F1 & AUC & F1 & AUC & F1 & AUC & F1 & AUC \\
\hline\hline
R50\cite{r} & 59.78 & 81.70 & 93.75 & 98.91 & 43.01 & 98.85 & 94.03 & 99.27 & 11.91 & 74.04 & 21.76 & 95.86 & 23.04 & 96.32 & 15.11 & 65.16  \\
R101\cite{r} & 71.93 & 83.64 & 94.84 & 99.21 & 39.22 & 96.98 & 85.36 & 97.62 & 11.20 & 73.30 & 32.77 & 96.24 & 22.31 & 93.87 & 24.22 & 99.20  \\
R152\cite{r} & 74.30 & \textit{87.49} & 91.97 & 98.57 & 43.04 & 98.18 & 70.21 & 87.90 & 10.67 & 71.37 & 32.42 & 96.58 & 19.22 & 83.09 & \textbf{24.58} & \textbf{99.61} \\
R50v2\cite{rv2} & \textit{78.96} & 87.32 & 96.60 & 99.44 & 47.99 & 99.79 & 92.82 & 99.06 & 11.42 & 73.11 & 34.11 & 96.32 & 23.93 & 98.72 & 18.71 & 78.27  \\
R101v2\cite{rv2} & 52.11 & 71.23 & 96.39 & 99.33 & 45.83 & 99.26 & 97.46 & 99.82 & 11.99 & 73.46 & 32.18 & 96.55 & 04.86 & 42.13 & 19.43 & 82.47  \\
R152v2\cite{rv2} & 53.44 & 71.97 & 95.30 & 99.01 & 47.10 & 99.71 & 97.76 & 99.82 & 11.93 & 73.23 & 32.69 & 96.54 & 23.07 & 95.89 & 23.03 & 96.25  \\
\hline
D121\cite{d} & 70.05 & 81.97 & 96.25 & 99.34 & 47.59 & 99.82 & 95.72 & 99.51 & 13.81 & 78.83 & 28.71 & 96.01 & 24.88 & 99.82 & 20.67 & 88.35   \\
D169\cite{d} & 59.18 & 76.56 & 88.86 & 95.60 & 46.40 & 99.68 & 94.33 & 99.52 & \textbf{15.21} & 79.90 & 32.90 & 96.46 & 24.97 & 99.84 & 20.13 & 85.95   \\
D201\cite{d} & 71.93 & 82.98 & 95.43 & 99.04 & \textit{48.17} & \textbf{99.83} & \textit{97.81} & 99.85 & 14.84 & 81.38 & 34.66 & 96.41 & \textbf{24.99} & \textbf{99.99} & 21.08 & 89.53   \\
\hline
ViT-B16\cite{vit} & 73.85 & 83.40 & 76.35 & 86.06 & 36.22 & 95.74 & 88.25 & 98.42 & 05.50 & 82.06 & 34.80 & 95.69 & 08.47 & 42.15 & 11.49 & 50.22  \\
ViT-B32\cite{vit} & 70.02 & 76.41 & 79.11 & 90.74 & 30.42 & 92.12 & 86.73 & 98.09 & 06.51 & 83.77 & 30.57 & 94.58 & 17.50 & 76.52 & 18.26 & 77.75  \\
ViT-L16\cite{vit} & 69.59 & 83.31 & 85.41 & 94.53 & 34.16 & 95.75 & 90.11 & 98.70 & 08.16 & 81.60 & 33.99 & 95.40 & 11.17 & 47.79 & 15.54 & 62.72  \\
ViT-L32\cite{vit} & 76.38 & 87.07 & 69.32 & 88.86 & 28.45 & 92.54 & 88.40 & 98.35 & 06.35 & 84.96 & 33.24 & 95.36 & 10.21 & 47.35 & 03.92 & 30.82  \\
CCT\cite{cct} & 62.10 & 71.18 & 80.60 & 92.04 & 32.63 & 95.33 & 92.52 & 99.11 & 08.08 & 85.37 & 30.25 & 95.12 & 23.98 & 98.53 & 19.43 & 83.21  \\
Swin0\cite{swin} & 66.04 & 83.74 & 96.27 & 99.57 & 47.63 & 99.66 & 97.53 & \textit{99.92} & 07.90 & 74.62 & 34.30 & 95.08 & 13.74 & 63.07 & 17.77 & 75.47  \\
Swin1\cite{swin} & 73.74 & 86.91 & \textit{96.65} & \textit{99.58} & 47.30 & 99.56 & 94.94 & 99.64 & 08.30 & 74.18 & 34.27 & 95.13 & 15.47 & 69.00 & 17.64 & 73.68  \\
\hline
\hline
%\textbf{RadT w/o (HVAT+VAL)} & \textbf{79.56} & \textbf{89.82} & 97.85 & 99.78 & 48.42 & 99.69 & 98.13 & 99.94 & 05.97 & \textbf{85.48} & \textbf{37.64} & 96.83 & 16.70 & 71.78 & 22.19 & 93.75  \\
\textbf{RadT} & \textbf{77.40} & \textbf{88.80} & \textbf{98.75} & \textbf{99.85} & \textbf{48.74} & 99.65 & \textbf{99.39} & \textbf{99.98} & 04.21 & \textbf{85.43} & \textbf{37.32} & \textbf{96.84} & 18.17 & 79.60 & 22.18 & 94.76  \\
\hline
\end{tabular}
}
\label{tab_2}
\end{center}
\end{table*}

\subsection{Quantitative results}
We report the F1 Score and Area-Under-Curve (AUC) for all experiments. Detailed results are shown in the Supplementary section. We compare our method with architectures such as different variations of ResNet~\cite{r}, ResNetv2~\cite{rv2}, DenseNet~\cite{d}, Vision Transformer~\cite{vit}, Compact Convolution Transformers~\cite{cct}, and two variations of Swin Transformers~\cite{swin}. Note that we show our comparison results primarily on the most prominent backbones (DenseNet-121~\cite{d}, vision transformer~\cite{vit}, etc.) used by the baselines~\cite{c11,c12,c27} and not on individual implementations. 
As shown in Table~\ref{tab_2}, our proposed architecture, mentioned as \textit{RadT}, outperforms other methods on all six datasets. Note that the F1 scores are computed without any standard averaging such as macro, micro or weighted. This is why, F1 scores on 14-class classification datasets, such as, NIH, and VinBigData are comparatively lower than the reported scores on RSNA, Radiography, etc. However, in these datasets where lower F1 scores are reported, the AUC of the proposed framework still outperforms the baselines.

\begin{table*}
%\tiny
\begin{center}
\caption{\textbf{Ablation Study.} Accuracy($\uparrow$), AUC($\uparrow$), F1($\uparrow$), Precision($\uparrow$), and Recall($\uparrow$) are shown for different ablations on three datasets}
\scalebox{0.67}{
\begin{tabular}{*{16}{c}}
\hline
Dataset$\rightarrow$ & \multicolumn{5}{c}{RSNA\cite{rsna}} & \multicolumn{5}{c}{Radiography\cite{rad_1,rad}} & \multicolumn{5}{c}{VinBigData\cite{vbd}} \\
\hline
Ablations$\downarrow$ & Ac. & AUC & F1 & Pr. & Re. & Ac. & AUC & F1 & Pr. & Re. & Ac. & AUC & F1 & Pr. & Re.\\
\hline\hline
Focal & 85.01 & 92.69 & 80.96 & 85.01 & 85.01 & 91.05 & 98.92 & 90.82 & 91.38 & 90.60 & 63.18 & 95.62 & 28.34 & 94.84 & 48.19  \\
Global & 86.45 & 93.99 & 83.26 & 86.45 & 86.45 & 89.91 & 98.65 & 88.90 & 90.38 & 89.44 & 62.46 & 95.46 & 25.79 & \textbf{95.53} & 47.54  \\
Focal+HVAT & 87.00 & 94.12 & 84.15 & 87.00 & 87.00 & 92.33 & 99.08 & 91.46 & 92.82 & 91.55 & 65.43 & 96.35 & 33.18 & 90.24 & 51.81  \\
Global+HVAT & 90.46 & 96.29 & 88.60 & 90.46 & 90.46 & 91.26 & 98.76 & 90.41 & 91.52 & 91.00 & 65.02 & 96.32 & 32.56 & 92.27 & 50.17  \\
Focal+HVAT+VAL & 89.68 & 95.88 & 87.62 & 89.68 & 89.68 & 93.04 & 99.22 & 92.66 & 93.35 & 92.66 & 65.32 & 96.31 & 33.49 & 92.30 & 50.44  \\
Global+HVAT+VAL & 89.76 & 96.00 & 87.51 & 89.76 & 89.76 & 91.05 & 98.76 & 90.32 & 91.47 & 90.60 & 64.97 & 96.16 & 31.85 & 91.73 & 50.41  \\
\hline
GF+HVAT+VAL(\textbf{RadT}) & \textbf{98.94} & \textbf{99.85} & \textbf{98.75} & \textbf{98.94} & \textbf{98.94} & \textbf{99.43} & \textbf{99.98} & \textbf{99.39} & \textbf{99.48} & \textbf{99.41} & \textbf{66.54} & \textbf{96.84} & \textbf{37.32} & 82.35 & \textbf{57.90}  \\
\hline
\end{tabular}
}
\label{tab_3}
\end{center}
\end{table*}

\textbf{Ablation experiments.} Here, we discuss the categorical inference on all the individual components of our proposed network. In Table~\ref{tab_3}, the ablation experiment results for different components are summarized for three different datasets. The global network outperforms the focal network for the binary classification task in the RSNA dataset. 
This signifies that for simple binary classification, where global feature representations generally lead to a clear distinction between labels, the global network performs better. This is, in fact, true for radiologists' decision making as well; the results provide a justification for the designed global-focal approach. For the Radiography and VinBigData datasets, which are multi-class classification tasks, focal network performs better than the global network owing to diagnostic relevance of the more granular details in the images. It is also evident from the results that when HVAT is used along with global-focal networks, the scores improve. Interestingly, when VAL is added, scores are not significantly higher than the previous ablations. There are primarily two reasons: a) VAL lacks in   
distilling the visual attention from the teacher to the student when using only individual global and focal blocks; the performance improves when VAL distills the visual attention from combined global-focal blocks of the teacher, and b) attention loss between the two visual attention regions may not converge well with regression of key-points and minimizing of GIoU.

\begin{figure}[t]
  \centering
  \includegraphics[width=0.6\linewidth]{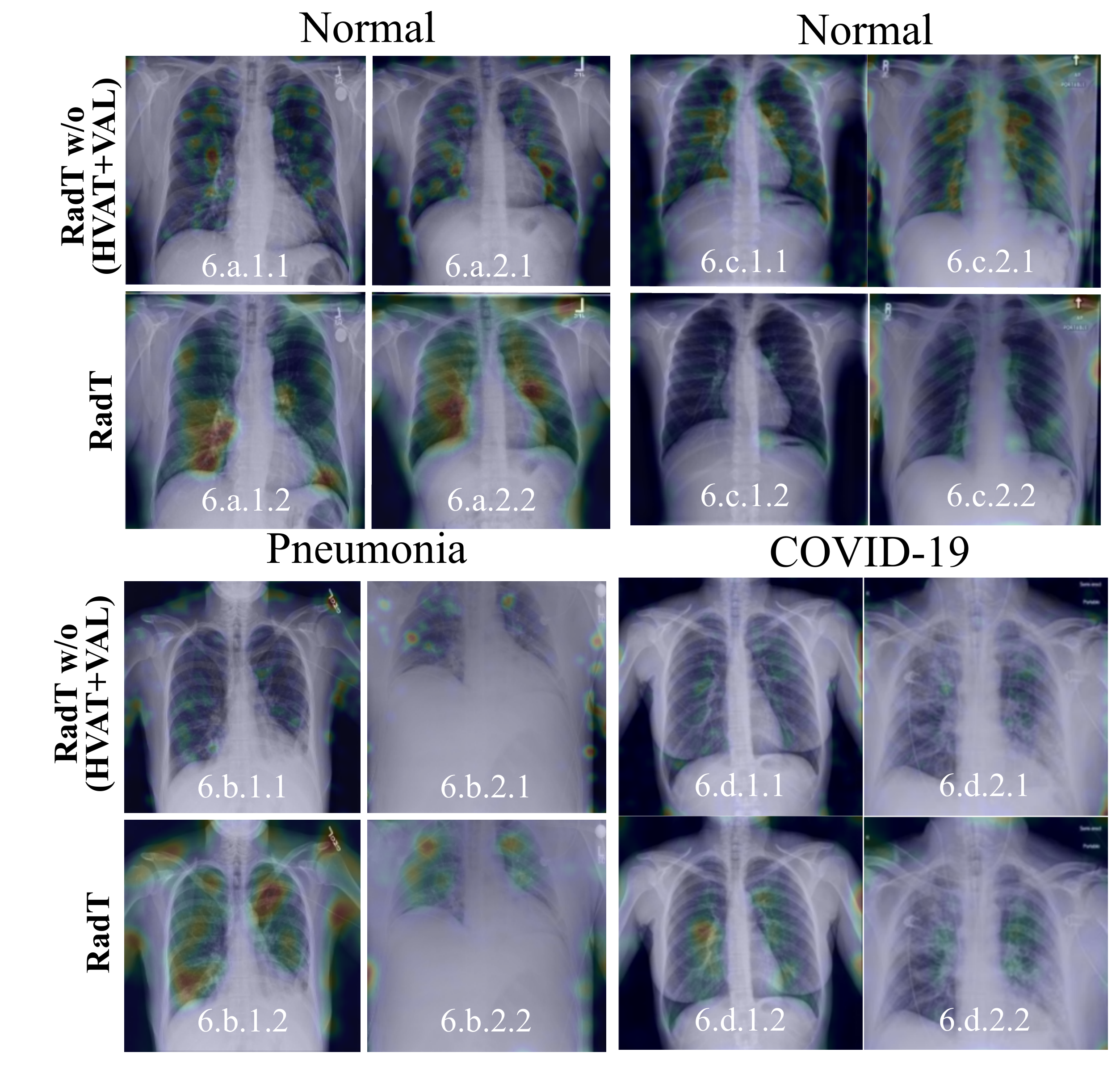}
   \caption{\textbf{Qualitative Comparison}. Comparison of the class activation maps from \systemname~ are shown on two datasets. \{6.a.*.* , 6.b.*.*\} are from \{normal, pneumonia\} classes of the RSNA dataset, and \{6.c.*.*, 6.d.*.*\} are from \{normal, COVID-19\} of the Radiography dataset. \{6.*.*.1, 6.*.*.2\} are the class activation maps generated from \{RadT w/o (HVAT+HVAL), RadT\}.}
   \label{fig_7}
\end{figure}

\subsection{Qualitative results}
Figure~\ref{fig_7} illustrates the qualitative differences between \textit{RadT w/o (HVAT+VAL)}, and \textit{RadT}. \textit{RadT w/o (HVAT+VAL)} is the basic backbone of our proposed \systemname~ architecture, i.e., the global-focal student-teacher network without HVAT and VAL. The first column, ~\ref{fig_7}.a.*.*, and~\ref{fig_7}.b.*.*, are normal and pneumonia samples from the RSNA dataset. Similarly,~\ref{fig_7}.c.*.* are normal, and~\ref{fig_7}.d.*.* are COVID-19, from the Radiography dataset. The images in~\ref{fig_7}.*.*.1 and \ref{fig_7}.*.*.2 are the class activation maps from RadT w/o (HVAT+VAL) and RadT, respectively. We can observe clear differences in attention region patterns between these two rows. The attention regions in the first row are relatively discretized and the inconsistency in overlap with the white regions (infiltrates/fluids) is quite prominent. However, in the second row, relatively continuous attention regions are observed with consistent overlap with the disease patterns. Similarly, in~\ref{fig_7}.c.*.1, attention regions observed are more discrete in nature, unlike~\ref{fig_7}.c.*.2. For normal chest radiographs, this potentially signifies that RadT focuses intrinsically on regions that may be significant for a radiologist to diagnose and reject the presence of infiltrates/fluids. On the contrary, RadT w/o (HVAT+VAL) attempts to identify non-overlapping regions with visual attention to reject the presence of infiltrates/fluids. Also, we observe that the attention regions from RadT w/o (HVAT+VAL) cover a larger area than those from RadT, implying that lack of visual attention knowledge leads to low confidence in decision-making and hence the model needs to search a comparatively larger space to conclusively accept or reject a claim. In~\ref{fig_7}.b.2.*, it is observed that for a lung densely filled with fluid, RadT w/o (HVAT+VAL) focuses on a comparatively sparse and large region. However, RadT focuses on regions with dense fluid accumulation. These qualitative findings suggest that \systemname~ inherently analyzes the regions with a visuo-cognitive approach similar to that of a radiologist.
\section{Conclusion}
This paper presents \systemname~, a novel visual attention--driven transformer framework, motivated by radiologists' visuo-cognitive approaches. Unlike existing techniques that rely only on visual information for diagnostic tasks, \systemname~ leverages eye-gaze patterns from experts to train a global-focal student-teacher network. 
Our framework learns and implements hierarchical search patterns to improve the diagnostic performance of transformer architectures.
When evaluated on eight datasets, comprising over 260,000 images, the proposed architecture outperforms SOTA approaches. Our qualitative analysis shows that by integrating visual attention into the network, \systemname~ focuses on diagnostically relevant regions of interest leading to higher confidence in decision making.
To the best of our knowledge, no method has been proposed that integrates gaze data from expert radiologists to improve the diagnostic performance of 
deep learning architectures. This work paves the way for radiologist-in-the-loop computer-aided diagnosis tools.

\textbf{Acknowledgements}
The reported research was partly supported by NIH 1R21CA258493-01A1, NIH
75N92020D00021 (subcontract), and the OVPR and IEDM seed grants at Stony
Brook University. The content is solely the responsibility of the authors and does not necessarily represent the official views of the National Institutes of Health.

\bibliographystyle{splncs04}
\bibliography{reference}
\end{document}

% \linenumbers
\pagestyle{headings}
\mainmatter
\def\ECCVSubNumber{7045}  % Insert your submission number here

\title{RadioTransformer: A Cascaded Global-Focal Transformer for Visual Attention--guided Disease Classification \\--- Supplementary Material ---} 
% \title{Supplementary Material for ``RadioTransformer: A Cascaded Global-Focal Transformer for Visual Attention--guided Disease Classification"} % Replace with your title

\begin{comment}
\titlerunning{ECCV-22 submission ID \ECCVSubNumber} 
\authorrunning{ECCV-22 submission ID \ECCVSubNumber} 
\author{Anonymous ECCV submission}
\institute{Paper ID \ECCVSubNumber}
\end{comment}
%******************

% \begin{comment}
\titlerunning{RadioTransformer}

\author{Moinak Bhattacharya
\orcidlink{0000-0003-2378-5632} \and
Shubham Jain
\orcidlink{0000-0002-4864-6420} \and
Prateek Prasanna
\orcidlink{0000-0002-3068-3573}}
\authorrunning{M. Bhattacharya et al.}
\institute{Stony Brook University, Stony Brook, New York, USA
\email{\{moinak.bhattacharya,prateek.prasanna\}@stonybrook.edu}}
% \end{comment}
%******************
\maketitle

\begin{figure*}[t]
  \centering
  \includegraphics[width=1.0\linewidth]{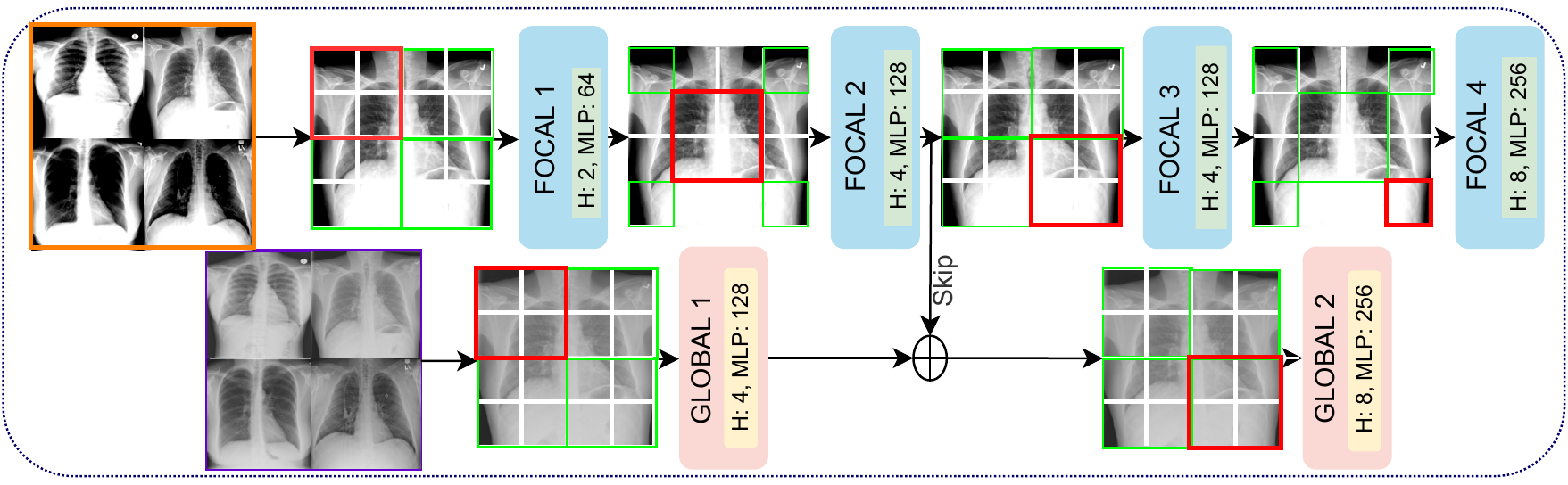}
   \caption{\textbf{Illustration of Global-Focal Network.} The Focal network (top row) learns low-level representations with high-contrast images as input, as shown in the orange box. The global network (bottom row) learns high-level representations with low-contrast images as input, as shown in the violet box. The shifting windows, shown as red boxes, are implemented with incremental shift size, shown as the traversing of the red boxes diagonally. For the global network, there is a single shifting, and for the focal network, there are three incremental shifting of the windows. Both the windows shift from top-left to bottom-right. The number of Attention heads (H) and MLP heads (MLP) for different global and focal blocks are also shown.}
   \label{fig_1}
\end{figure*}

In this supplementary material, we provide detailed illustration of the global-focal block (Section~\ref{gfb}), additional information on the datasets used in this work (Section~\ref{assets}), the different augmentations in student-teacher network (Section~\ref{aug}), more quantitative (Section~\ref{quan_ana}), and qualitative (Section~\ref{qual_ana}) results. We also present an analogy of the global-focal block with cellular pathways (Section~\ref{ana_cp}).
\section{Illustration of Global-Focal block}
\label{gfb}
The global-focal block in the RadioTransformer architecture is detailed in Figure~\ref{fig_1}. The global and focal blocks are cascaded in parallel. The shifting window for each block is shown with the window in red color. High contrast patterns are learned by the focal blocks, shown in the orange box in Figure~\ref{fig_1} and low contrast patterns are learned by global blocks, shown in the blue box in Figure~\ref{fig_1}. The TWL connection averages the features between the intermediate global and focal blocks.

\section{Datasets}
\label{assets}
RSNA Pneumonia Detection challenge\cite{rsna}, and Cell Pneumonia\cite{cell} are pneumonia classification datasets consisting of radiographs with presence and absence of pneumonia. SIIM-FISABIO-RSNA COVID-19 Detection\cite{siim} dataset categorizes radiographs as negative for pneumonia, and typical, indeterminate, or atypical for COVID-19. COVID-19 Radiography database\cite{rad,rad_1} comprises chest radiographs with COVID-19, normal, lung opacity and viral pneumonia classes. NIH Chest X-rays\cite{nih} and VinBigData Chest X-ray Abnormalities Detection\cite{vbd} datasets comprise 14 common thorax diseases. We further include the more recent large-scale RSNA-MIDRC\cite{midrc,midrc_1,clark2013cancer} and TCIA-SBU COVID-19 datasets~\cite{sbu,clark2013cancer} that contain only COVID-19 chest radiographs. 

\section{Augmentation}
\label{aug}
\begin{figure*}[t]
  \centering
  \includegraphics[width=1.0\linewidth]{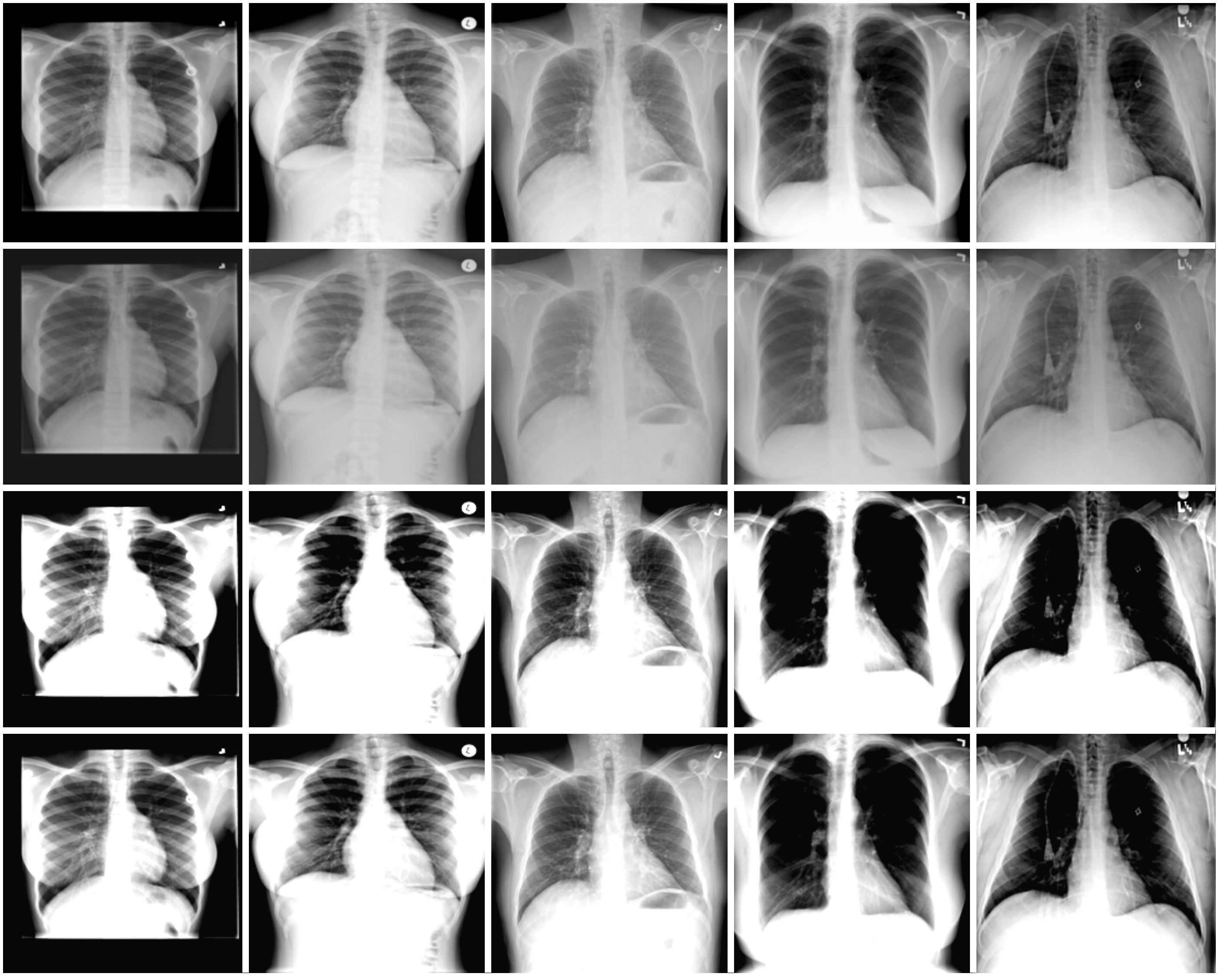}
   \caption{\textbf{Augmentations}: The different augmentations to input images of student global-focal, and teacher global-focal blocks are shown.}
   \label{fig_aug}
\end{figure*}

Figure~\ref{fig_aug} illustrates the various augmentations for different blocks of \textit{RadioTransformer}. The images in the first and second rows are the inputs to the student focal and global blocks, respectively. The images in the third and fourth rows are the inputs to teacher focal and global blocks, respectively. As seen in the images, the teacher network implements hard augmentations compared to the student network. The focal block has a higher contrast value than the global block. For stateless augmentations, we use tf.image.stateless\_random\_contrast(.), tf.image.stateless\_random\_brightness(.), tf.image.stateless-\\\_random\_hue(.), and tf.image.stateless\_random\_saturation(.). More details on the augmentation parameters are provided in Supplementary table~\ref{tab_aug_supp}.

\begin{table}
\begin{center}
\scalebox{0.8}{
\begin{tabular}{*{7}{c}}
\hline
Augmentation & \multicolumn{2}{c}{Contrast} & Brightness & Hue & \multicolumn{2}{c}{Saturation}\\
\hline
Parameter & lower & upper & max\_delta & max\_delta & lower & upper\\
\hline
\hline
Teacher-Global & 2.0 & 2.2 & 0.8 & 0.8 & 2.0 & 2.5 \\
Teacher-Focal & 2.8 & 3.0 & 0.8 & 0.8 & 2.0 & 2.5 \\
Student-Global & 0.5 & 1.0 & 0.5 & 0.5 & 1.5 & 2.0 \\ 
Student-Focal & 1.0 & 1.5 & 0.5 & 0.5 & 1.5 & 2.0 \\
\hline
\end{tabular}
}
\end{center}
\caption{Augmentation parameters.}
\label{tab_aug_supp}
\end{table}

\begin{sidewaystable}
\begin{center}
\scalebox{0.75}{
\begin{tabular}{*{21}{c}}
\hline
Name & \multicolumn{5}{c}{Cell\cite{cell}} & \multicolumn{5}{c}{RSNA\cite{rsna}} & \multicolumn{5}{c}{SIIM\cite{siim}} & \multicolumn{5}{c}{Rad\cite{rad,rad_1}} \\
\hline
- & Acc. & AUC & F1 & Pr. & Re. & Acc. & AUC & F1 & Pr. & Re. & Acc. & AUC & F1 & Pr. & Re. & Acc. & AUC & F1 & Pr. & Re. \\
\hline\hline
R50\cite{r} & 71.35 & 81.70 & 59.78 & 68.34 & 73.44 & 94.56 & 98.91 & 93.75 & 94.78 & 94.15 & 89.90 & 98.85 & 43.01 & 89.90 & 89.90 & 94.41 & 99.27 & 94.03 & 94.54 & 94.25  \\
R101\cite{r} & 78.47 & 83.64 & 71.93 & 78.73 & 77.78 & 95.66 & 99.21 & 94.84 & 95.56 & 95.83 & 80.29 & 96.98 & 39.22 & 80.29 & 80.29 & 87.97 & 97.62 & 85.36 & 88.31 & 87.78
   \\
R152\cite{r} & 79.86 & 87.49 & 74.30 & 80.81 & 79.69 & 93.57 & 98.57 & 91.97 & 93.23 & 94.02 & 88.62 & 98.18 & 43.04 & 88.62 & 88.62 & 66.62 & 87.90 & 70.21 & 67.20 & 66.03
 \\
R50v2\cite{rv2} & 82.53 & 87.32 & 78.96 & 82.53 & 82.53 & 97.12 & 99.44 & 96.60 & 97.12 & 97.12 & 96.79 & 99.79 & 47.99 & 96.79 & 96.79 & 94.22 & 99.06 & 92.82 & 94.44 & 94.18
    \\
R101v2\cite{rv2} & 68.40 & 71.23 & 52.11 & 68.40 & 68.40 & 97.01 & 99.33 & 96.39 & 97.01 & 97.01 & 93.67 & 99.26 & 45.83 & 93.67 & 93.67 & 97.85 & 99.82 & 97.46 & 97.85 & 97.85
   \\
R152v2\cite{rv2} & 69.10 & 71.97 & 53.44 & 69.10 & 69.10 & 96.08 & 99.01 & 95.30 & 96.08 & 96.08 & 95.31 & 99.71 & 47.10 & 95.31 & 95.31 & 98.30 & 99.82 & 97.76 & 98.30 & 98.30
   \\
\hline
D121\cite{d} & 77.43 & 81.97 & 70.05 & 77.43 & 77.43 & 96.84 & 99.34 & 96.25 & 96.84 & 96.84 & 96.22 & 99.82 & 47.59 & 96.22 & 96.22 & 96.52 & 99.51 & 95.72 & 96.65 & 96.45
   \\
D169\cite{d} & 71.70 & 76.56 & 59.18 & 71.70 & 71.70 & 89.96 & 95.60 & 88.86 & 89.96 & 89.96 & 94.49 & 99.68 & 46.40 & 94.49 & 94.49 & 95.48 & 99.52 & 94.33 & 95.63 & 95.43
  \\
D201\cite{d} & 78.47 & 82.98 & 71.93 & 78.47 & 78.47 & 96.23 & 99.04 & 95.43 & 96.23 & 96.23 & 97.12 & 99.83 & 48.17 & 97.12 & 97.04 & 97.80 & 99.85 & 97.81 & 97.82 & 97.77
   \\
\hline
ViT-B16\cite{vit} & 76.91 & 83.40 & 73.85 & 76.91 & 76.91 & 78.08 & 86.06 & 76.35 & 78.08 & 78.08 & 78.45 & 95.74 & 36.22 & 78.53 & 78.21 & 89.51 & 98.42 & 88.25 & 90.04 & 89.02
 \\
ViT-B32\cite{vit} & 70.41 & 76.41 & 70.02 & 70.14 & 70.14 & 82.83 & 90.74 & 79.11 & 82.83 & 82.83 & 68.91 & 92.12 & 30.42 & 69.22 & 68.26 & 88.40 & 98.09 & 86.73 & 89.15 & 87.76
   \\
ViT-L16\cite{vit} & 75.69 & 83.31 & 69.59 & 75.69 & 75.69 & 87.85 & 94.53 & 85.41 & 87.85 & 87.85 & 78.29 & 95.75 & 34.16 & 78.29 & 78.29 & 90.91 & 98.70 & 90.11 & 91.36 & 90.67
   \\
ViT-L32\cite{vit} & 80.38 & 87.07 & 76.38 & 80.38 & 80.38 & 79.24 & 88.86 & 69.32 & 79.24 & 79.24 & 70.07 & 92.54 & 28.45 & 70.31 & 69.90 & 89.44 & 98.35 & 88.40 & 89.94 & 88.85
  \\
CCT\cite{cct} & 71.18 & 74.59 & 62.10 & 71.18 & 71.18 & 83.84 & 92.04 & 80.60 & 83.84 & 83.84 & 78.12 & 95.33 & 32.63 & 78.12 & 78.12 & 92.19 & 99.11 & 92.52 & 92.33 & 92.09
   \\
Swin0\cite{swin} & 74.83 & 83.74 & 66.04 & 75.13 & 73.96 & 96.87 & 99.57 & 96.27 & 96.79 & 97.11 & 96.38 & 99.66 & 47.63 & 72.19 & 99.92 & 97.94 & 99.92 & 97.53 & 98.31 & 97.54
  \\
Swin1\cite{swin} & 78.65 & 86.91 & 73.74 & 78.25 & 79.34 & 97.17 & 99.58 & 96.65 & 97.14 & 97.22 & 95.72 & 99.56 & 47.30 & 66.48 & 99.67 & 95.48 & 99.64 & 94.94 & 95.71 & 95.17
   \\
\hline
\hline
\textbf{RadT w/o (HVAT+VAL)} & 82.05 & 89.82 & 79.56 & 82.05 & 82.05 & 98.19 & 99.78 & 97.85 & 98.19 & 98.19 & 97.47 & 99.69 & 48.42 & 97.47 & 97.47 & 98.51 & 99.94 & 98.13 & 98.58 & 98.51
  \\
\textbf{RadT} & 80.73 & 88.80 & 77.40 & 77.65 & 82.64 & 98.94 & 99.85 & 98.75 & 98.94 & 98.94 & 98.10 & 99.65 & 48.74 & 98.10 & 98.10 & 99.43 & 99.98 & 99.39 & 99.48 & 99.41
   \\
\hline
\end{tabular}
}
\end{center}
\caption{Quantitative Comparison:1. F1($\uparrow$) and AUC($\uparrow$) are reported for the baselines and the proposed methodology.}
\label{tab_1_supp}
\end{sidewaystable}

\begin{sidewaystable}
\begin{center}
\scalebox{0.8}{
\begin{tabular}{*{21}{c}}
\hline
Name & \multicolumn{5}{c}{NIH\cite{nih}} & \multicolumn{5}{c}{VBD\cite{vbd}} & \multicolumn{5}{c}{MIDRC\cite{midrc_1,midrc}} & \multicolumn{5}{c}{SBU\cite{sbu,clark2013cancer}}\\
\hline
- & Acc. & AUC & F1 & Pr. & Re. & Acc. & AUC & F1 & Pr. & Re. & Acc. & AUC & F1 & Pr. & Re. & Acc. & AUC & F1 & Pr. & Re. \\
\hline\hline
R50\cite{r} & 35.28 & 74.04 & 11.91 & 36.03 & 34.38 & 62.96 & 95.86 & 21.76 & 97.11 & 48.50 & 85.44 & 96.32 & 23.04 & 85.83 & 85.20 & 43.29 & 65.16 & 15.11 & 43.29 & 42.93
  \\
R101\cite{r} & 32.95 & 73.30 & 11.20 & 33.81 & 31.98 & 66.41 & 96.24 & 32.77 & 88.45 & 55.03 & 80.59 & 93.87 & 22.31 & 80.88 & 80.35 & 93.92 & 99.20 & 24.22 & 94.37 & 93.52
   \\
R152\cite{r} & 30.13 & 71.37 & 10.67 & 30.51 & 29.28 & 66.47 & 96.58 & 32.42 & 90.62 & 53.24 & 62.42 & 83.09 & 19.22 & 62.78 & 62.01 & 96.73 & 99.61 & 24.58 & 96.93 & 96.54
\\
R50v2\cite{rv2} & 35.04 & 73.11 & 11.42 & 35.56 & 34.30 & 66.02 & 96.32 & 34.11 & 90.27 & 53.04 & 91.78 & 98.72 & 23.93 & 91.85 & 91.69 & 59.81 & 78.27 & 18.71 & 59.85 & 59.71
  \\
R101v2\cite{rv2} & 36.37 & 73.46 & 11.99 & 37.02 & 35.80 & 66.26 & 96.55 & 32.18 & 91.18 & 53.27 & 10.77 & 42.13 & 04.86 & 10.55 & 10.44 & 63.55 & 82.47 & 19.43 & 64.35 & 62.75
    \\
R152v2\cite{rv2} & 32.05 & 73.23 & 11.93 & 32.94 & 31.04 & 66.04 & 96.54 & 32.69 & 91.76 & 53.27 & 85.69 & 95.89 & 23.07 & 85.75 & 85.61 & 85.39 & 96.25 & 23.03 & 85.50 & 85.21
   \\
\hline
D121\cite{d} & 33.67 & 78.83 & 13.81 & 15.34 & 73.37 & 64.31 & 96.01 & 28.71 & 91.02 & 50.63 & 99.01 & 99.82 & 24.88 & 64.85 & 100.00 & 70.47 & 88.35 & 20.67 & 70.68 & 70.24
  \\
D169\cite{d} & 33.27 & 79.90 & 15.21 & 16.60 & 73.67 & 66.04 & 96.46 & 32.90 & 90.25 & 53.77 & 99.75 & 99.84 & 24.97 & 56.30 & 100.00 & 67.38 & 85.95 & 20.13 & 67.49 & 67.22
  \\
D201\cite{d} & 36.02 & 81.38 & 14.84 & 17.76 & 75.45 & 66.42 & 96.41 & 34.66 & 88.69 & 55.09 & 99.92 & 99.99 & 24.99 & 65.59 & 100.00 & 72.92 & 89.53 & 21.08 & 73.06 & 72.71
   \\
\hline
ViT-B16\cite{vit} & 34.54 & 82.06 & 07.50 & 42.99 & 21.24 & 64.14 & 95.69 & 34.80 & 83.00 & 54.67 & 20.39 & 42.15 & 08.47 & 19.93 & 19.33 & 29.84 & 50.22 & 11.49 & 28.85 & 26.41
 \\
ViT-B32\cite{vit} & 38.19 & 83.77 & 06.51 & 48.48 & 22.94 & 60.75 & 94.58 & 30.57 & 88.86 & 47.69 & 53.87 & 76.52 & 17.50 & 54.97 & 51.81 & 57.52 & 77.75 & 18.26 & 58.42 & 56.19
   \\
ViT-L16\cite{vit} & 32.22 & 81.60 & 08.16 & 43.32 & 16.28 & 63.66 & 95.40 & 33.99 & 80.80 & 55.15 & 28.78 & 47.79 & 11.17 & 28.52 & 27.47 & 45.08 & 62.72 & 15.54 & 45.38 & 43.69
   \\
ViT-L32\cite{vit} & 38.66 & 84.96 & 06.35 & 47.04 & 25.42 & 63.44 & 95.36 & 33.24 & 86.29 & 52.51 & 25.66 & 47.35 & 10.21 & 24.87 & 23.68 & 08.52 & 30.82 & 03.92 & 06.01 & 05.41
  \\
CCT\cite{cct} & 38.69 & 85.37 & 08.08 & 52.10 & 22.06 & 62.02 & 95.12 & 30.25 & 89.69 & 49.51 & 92.19 & 98.53 & 23.98 & 92.93 & 91.94 & 63.57 & 83.21 & 19.43 & 64.20 & 62.84
   \\
Swin0\cite{swin} & 31.65 & 74.62 & 07.90 & 33.21 & 28.14 & 64.81 & 95.08 & 34.30 & 16.57 & 97.32 & 37.91 & 63.07 & 13.74 & 37.96 & 37.34 & 55.12 & 75.47 & 17.77 & 55.42 & 54.43
  \\
Swin1\cite{swin} & 31.17 & 74.18 & 08.30 & 32.71 & 27.34 & 65.03 & 95.13 & 34.27 & 16.48 & 97.83 & 44.82 & 69.00 & 15.47 & 45.10 & 44.24 & 54.50 & 73.68 & 17.64 & 54.77 & 53.99
   \\
\hline
\hline
\textbf{RadT w/o (HVAT+VAL)} & 38.56 & 85.48 & 05.97 & 49.92 & 20.33 & 65.96 & 96.83 & 37.64 & 83.87 & 56.49 & 50.17 & 71.78 & 16.70 & 50.47 & 49.67 & 79.79 & 93.75 & 22.19 & 81.84 & 77.71
  \\
\textbf{RadT} & 38.52 & 85.43 & 04.21 & 45.48 & 26.73 & 66.54 & 96.84 & 37.32 & 82.35 & 57.90 & 57.07 & 79.60 & 18.17 & 43.48 & 72.45 & 79.69 & 94.76 & 22.18 & 83.89 & 75.15
   \\
\hline
\end{tabular}
}
\end{center}
\caption{Quantitative Comparison:2. F1($\uparrow$) and AUC($\uparrow$) are reported for the baselines and the proposed methodology.}
\label{tab_2_supp}
\end{sidewaystable}

\begin{figure*}[t]
  \centering
  \includegraphics[width=1.0\linewidth]{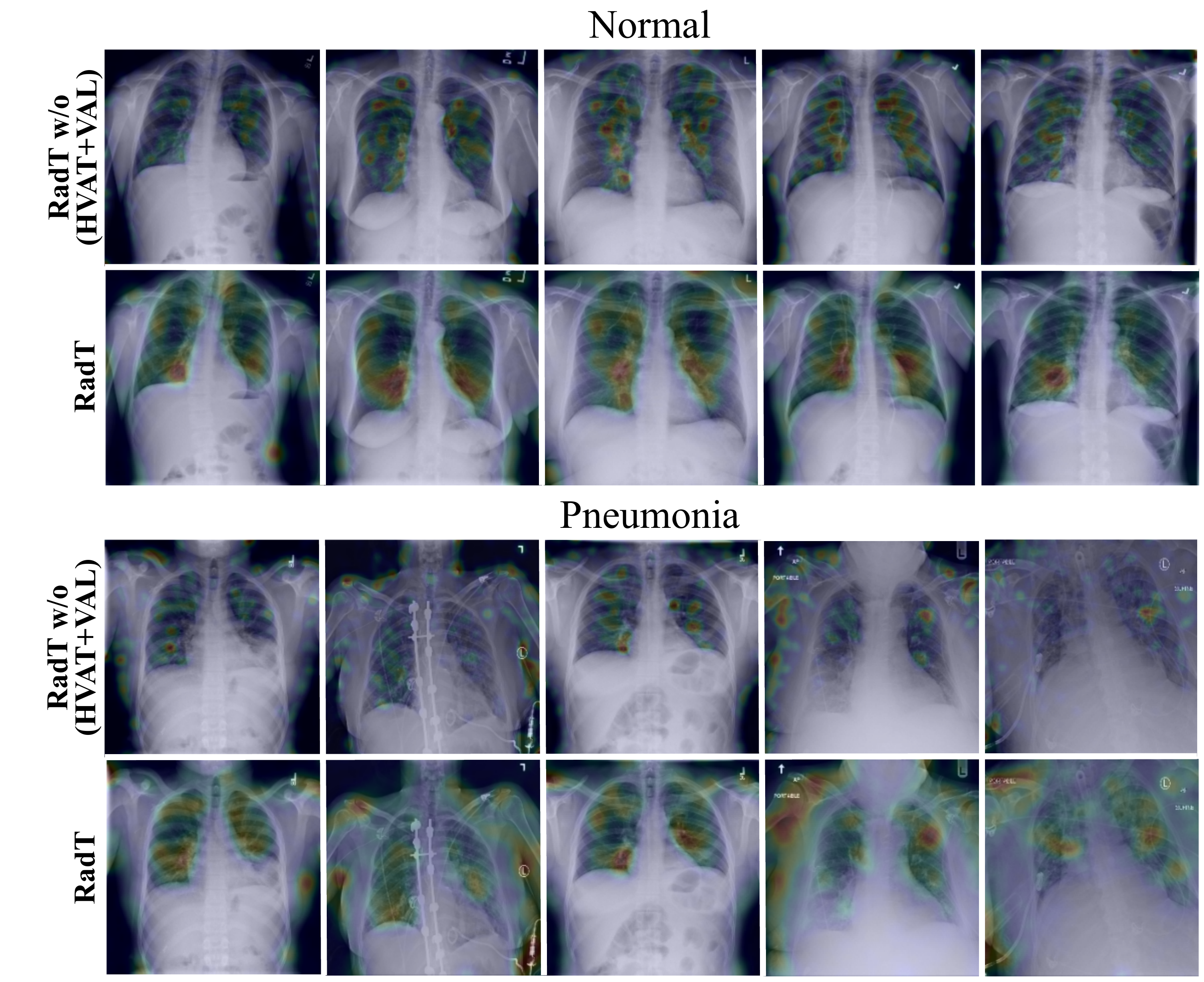}
   \caption{\textbf{Qualitative Comparison on RSNA dataset:} Comparison of class activation maps from RadioTransformer w/o (HVAT+VAL) and RadioTransformer.}
   \label{fig_rsna_cam}
\end{figure*}

\begin{figure*}[t]
  \centering
  \includegraphics[width=1.0\linewidth]{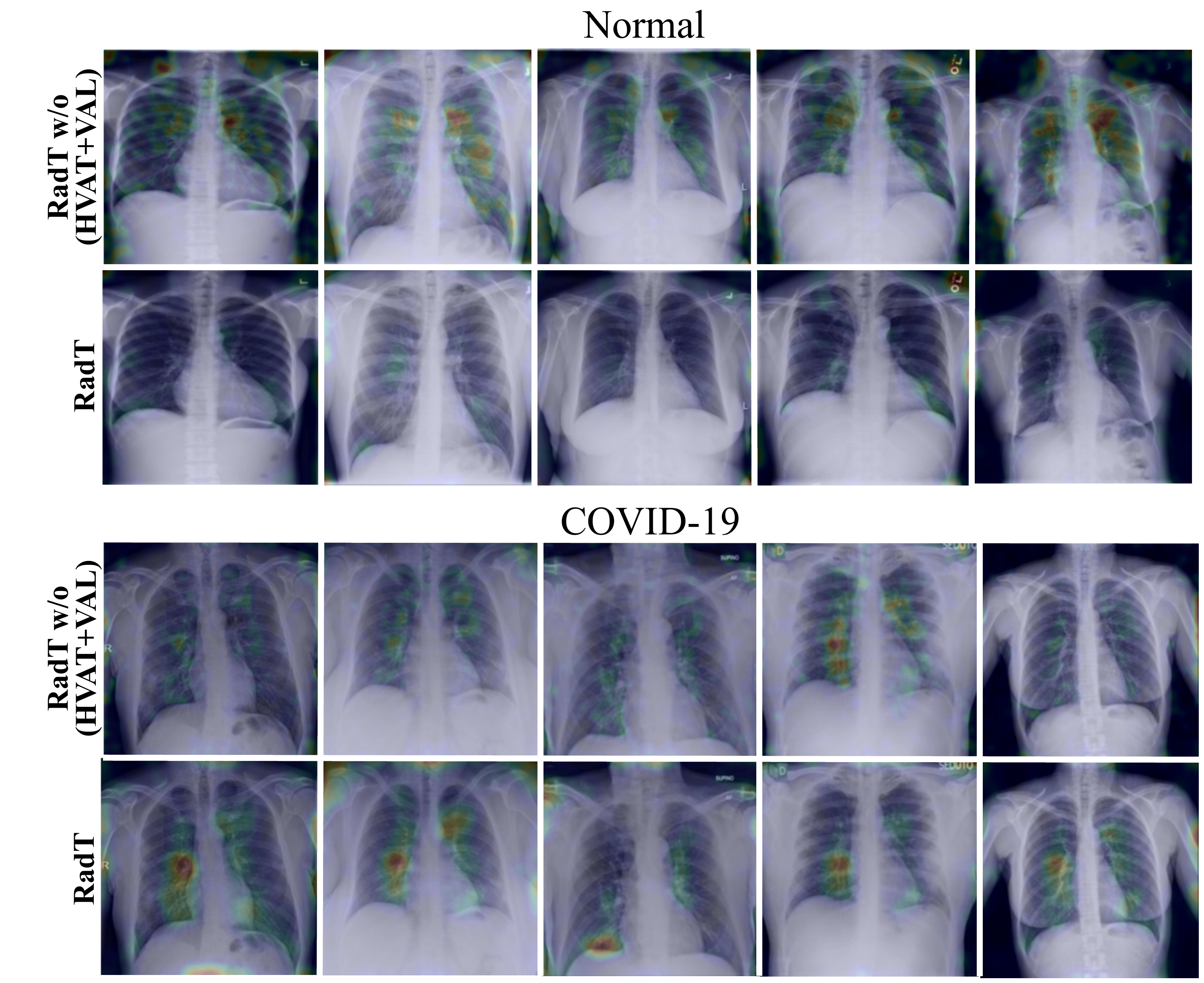}
   \caption{\textbf{Qualitative Comparison on Radiography dataset:} Comparison of class activation maps from RadioTransformer w/o (HVAT+VAL) and RadioTransformer.}
   \label{fig_rad_cam}
\end{figure*}

\section{Quantitative Analysis}
\label{quan_ana}
In addition to the AUC and F1 scores provided in the main paper, here we show the accuracy, precision, and recall values for classification tasks in the 8 datasets. 
In Supplementary table~\ref{tab_1_supp}, the performance metrics for pneumonia classification datasets such as Cell, and RSNA Pneumonia Challenge dataset, and COVID-19 classification datasets such as SIIM-RSNA-FISABIO COVID-19 challenge, and Radiography dataset are shown. In Supplementary table~\ref{tab_2_supp}, we show the performance metrics for 14 thoracic diseases classification tasks (in the NIH, and VinBigData datasets), and the COVID-19 classification task (in MIDRC and TCIA-SBU datasets). 
\section{Qualitative Analysis}
\label{qual_ana}
We supplement our qualitative results (in Section 5.2 of the main paper) with additional  class activation maps for both the datasets i.e., RSNA, and Radiography. 
In Figure~\ref{fig_rsna_cam}, the RadT w/o (HVAT+VAL) and RadT class activation maps are shown for Normal and Pneumonia cases. Similarly, in Figure ~\ref{fig_rad_cam}, the RadT w/o (HVAT+VAL), and RadT class activation maps are shown for Normal and COVID-19 cases. For both the datasets, the maps of RadT w/o (HVAT+VAL) show discrete patterns, and those of RadT show comparatively continuous patterns. In addition to all the previous discussions, we discuss another interesting finding. In the fourth row of Figure~\ref{fig_rsna_cam}, we observe that apart from clear attention on the white/fluid regions, there are some extraneous attention regions in the shoulders. Again, this phenomenon is not observed in the fourth row of Figure~\ref{fig_rad_cam}. This is clearly explainable from the ablation study in the main paper. For the RSNA dataset, the global block is showing better performance, hence the global block is activated in this case. The global block focuses on high-level features and in this case, it hypothesizes to identify features from non-relevant regions(like shoulder, etc) in addition to the white/fluid regions in the lungs. Whereas in the Radiography dataset, the focal block is activated and the attention regions perfectly intersect with the white/fluid regions.  

\section{Analogy with cellular pathways}
\label{ana_cp} 
Parvo, Magno, and Konio cells are ganglion cells that transfer information generated by the photoreceptors in the retina to the visual cortex in the brain.  Structurally,  Magno cells are larger, and have thick axons with more myelin while Parvo cells are smaller, and have less myelin and thinner axons. Functionally, the Magno cells have a large receptive field; they respond rapidly to changing stimuli and detect robust/global details like luminance, motion, stereopsis, and depth. Parvo cells, on the other hand, have a smaller receptive field, respond slowly to stimuli, and detect finer/local details like chromatic modulation and the form of an object. The Global-Focal blocks in \textit{RadioTransformer} are inspired by these cellular pathways.

\bibliographystyle{splncs04}
\bibliography{egbib}